\documentclass[10pt,twocolumn,letterpaper]{article}

 \usepackage[pagenumbers]{cvpr} 

\definecolor{cvprblue}{rgb}{0.21,0.49,0.74}
\usepackage[pagebackref,breaklinks,colorlinks,allcolors=cvprblue]{hyperref}

\def\paperID{8306} 
\def\confName{CVPR}
\def\confYear{2026}

\title{coDrawAgents: A Multi-Agent Dialogue Framework for Compositional Image Generation}

\author{
Chunhan Li$^{1,7}$ \quad Qifeng Wu$^{2}$ \quad Jia-Hui Pan$^3$ \quad Ka-Hei Hui$^3$ \quad Jingyu Hu$^3$ \quad Yuming Jiang$^4$ \\
Bin Sheng$^5$ \quad Xihui Liu$^6$ \quad Wenjuan Gong$^7$ \quad Zhengzhe Liu$^{1}$\thanks{Corresponding author: {\tt zhengzheliu@ln.edu.hk}}
\and
$^1$Lingnan University. \quad $^2$CMU \quad $^3$CUHK \quad $^4$Alibaba DAMO Academy \\
$^5$SJTU \quad $^6$HKU \quad $^7$China University of Petroleum
}

\begin{document}
\maketitle
\begin{abstract}
Text-to-image generation has advanced rapidly, but existing models still struggle with faithfully composing multiple objects and preserving their attributes in complex scenes. We propose \textbf{coDrawAgents}, an interactive multi-agent dialogue framework with four specialized agents: Interpreter, Planner, Checker, and Painter that collaborate to improve compositional generation. The Interpreter adaptively decides between a direct text-to-image pathway and a layout-aware multi-agent process. In the layout-aware mode, it parses the prompt into attribute-rich object descriptors, ranks them by semantic salience, and groups objects with the same semantic priority level for joint generation. 
Guided by the Interpreter, the Planner adopts a divide-and-conquer strategy, incrementally proposing layouts for objects with the same semantic priority level while grounding decisions in the evolving visual context of the canvas.
The Checker introduces an explicit error-correction mechanism by validating spatial consistency and attribute alignment, and refining layouts before they are rendered. 
Finally, the Painter synthesizes the image step by step, incorporating newly planned objects into the canvas to provide richer context for subsequent iterations. Together, these agents address three key challenges: reducing layout complexity, grounding planning in visual context, and enabling explicit error correction. Extensive experiments on benchmarks GenEval and DPG-Bench demonstrate that coDrawAgents substantially improves text–image alignment, spatial accuracy, and attribute binding compared to existing methods.

\end{abstract}

\vspace{-13.5pt}
\section{Introduction}

Text-to-Image (T2I) generation has emerged as a pivotal area in artificial intelligence, enabling the creation of visual content from textual descriptions \citep{rombach2022high}. However, current T2I models often struggle with user controllability, particularly concerning the reasonable arrangement and relationships of objects within the generated image \citep{zhang2023adding, hertz2022prompt}. These models can exhibit numerical and spatial inaccuracies, leading to challenges in tasks such as faithful layout arrangement \citep{zheng2023layout} and maintaining compositional faithfulness, where the generated image accurately reflects the structure and relationships described in the text \citep{chefer2023attend, liu2022compositional}.

To address the above limitations, existing works have explored the use of Large Language Models (LLMs) and agentic frameworks to assist in the generation of spatial layouts for images. Some approaches \citep{feng2023layoutgpt,lian2023llm} leverage the planning capabilities of \textbf{LLMs} to interpret the input text and propose arrangements for the described objects, aiming to improve the structural coherence of the generated images.  
Furthermore, recent studies explore \textbf{agent-based frameworks} that uses multiple specialized LLM agents for text-to-image generation.
For instance, MCCD \citep{li2025mccd} demonstrates the effectiveness of multi-agent collaboration for compositional diffusion.
However, both single-agent and existing multi-agent systems lack the interactive, closed-loop reasoning required for reliable compositional generation.
Single-agent methods place all parsing, planning, and verification on one model, making early spatial mistakes difficult to detect or revise.
Meanwhile, many multi-agent frameworks are essentially fixed pipelines without negotiation or visual grounding, so errors still propagate uncorrected.
As a result, neither paradigm provides the mutual checking, iterative refinement, or canvas-aware reasoning needed for stable layout planning.

These limitations are further amplified in scenes containing multiple objects, where stable layout planning requires overcoming several inherent challenges. 
First, global layout planning incurs quadratic relational complexity among objects, making it difficult for a single planner to capture all dependencies~\citep{zheng2023layout, feng2023layoutgpt}. 
Second, most approaches predict layouts without access to visual context, forcing the planner to “imagine” the scene in isolation, which often leads to incoherent or unrealistic arrangements. 
%
Third, most existing works use diffusion-based models and diffusion pipelines typically commit to a coarse global structure in early denoising steps, with fine details added only later \citep{hertz2022prompt, chefer2023attend}. As a result, errors such as misplaced objects or incorrect attributes, once introduced early, are difficult to correct due to the lack of explicit error-correction mechanisms. 

In this work, we propose an interactive multi-agent dialogue framework with four specialized agents: an Interpreter for generative mode selection and text decomposition, a Planner for incremental layout reasoning, a Checker for spatial and semantic verification, and a Painter for visual synthesis. 
The Interpreter determines whether to invoke the layout-free mode that connect the Painter directly or to activate the layout-aware mode for complex scenes. 
In the layout-free mode, the interpreter directly call a text-to-image painter for generation. On the contrary, in the layout-aware mode, unlike pipeline-based designs, our four agents engage in dynamic dialogue: it parses the text into attribute enriched object descriptions, ranks them by semantic salience, groups equal-priority objects for joint generation, and schedules iterative plans. Then the Planner incrementally proposes layouts one object (or group) at a time, the Checker validates spatial and semantic consistency against the text and evolving scene, and the Painter synthesizes the image step by step in a training-free and plug-and-play manner, with the evolving canvas providing crucial visual context for subsequent iterations.

Our framework effectively addresses the core challenges of prior methods. 
First, instead of performing global planning over all objects simultaneously, the Planner, guided by the Interpreter, adopts a divide-and-conquer strategy by reasoning about objects 
with the same semantic priority level
at a time, thereby substantially reducing layout complexity.
Second, the Planner leverages the evolving visual context from the Painter’s canvas, ensuring that layout predictions are grounded in the actual scene rather than imagined in isolation. 
Third, the Checker introduces an explicit error-correction mechanism, validating object placement alignment, and applying necessary adjustments to improve layout faithfulness.
Unlike single-agent or pipeline systems, our agents reason collaboratively, allowing planning, checking, and synthesis to inform one another. This closed-loop interaction enables codrawAgents to revise errors, yielding more stable and faithful alignment between text and image.


We evaluate coDrawAgents on the GenEval and DPG-Bench benchmarks. On GenEval, our framework sets a new state of the art with substantial improvements in compositional fidelity, object relations, and attribute binding over prior approaches. On DPG-Bench, which stresses long-context and multi-object reasoning, coDrawAgents consistently outperforms recent baselines, achieving stronger spatial accuracy and text–image consistency. Qualitative comparisons further highlight these gains: whereas existing methods often misplace objects, miscount, or confuse attributes, coDrawAgents generates coherent layouts and high-quality images aligned with the textual descriptions. These results demonstrate the effectiveness of our interactive multi-agent design for compositional image generation tasks. \textbf{Codes will be released upon publication.}

In summary, our contributions are threefold:
\begin{itemize}
 \item We introduce \textbf{coDrawAgents}, an interactive multi-agent dialogue framework with four specialized agents: Interpreter, Planner, Checker, and Painter that collaborate dynamically instead of following a fixed pipeline.
 \item We propose three technical innovations: (i) a divide-and-conquer planning strategy that reduces layout complexity, (ii) grounding layout decisions in the evolving visual context for stronger spatial alignment, and (iii) an explicit error-correction mechanism via the Checker to enhance faithfulness. 
 \item We achieve \textbf{state-of-the-art performance} on GenEval and DPG-Bench, showing clear gains in text–image consistency, spatial accuracy, and attribute binding.
\end{itemize}

\begin{figure*}[t]
\begin{center}
\includegraphics[width=0.85\textwidth]{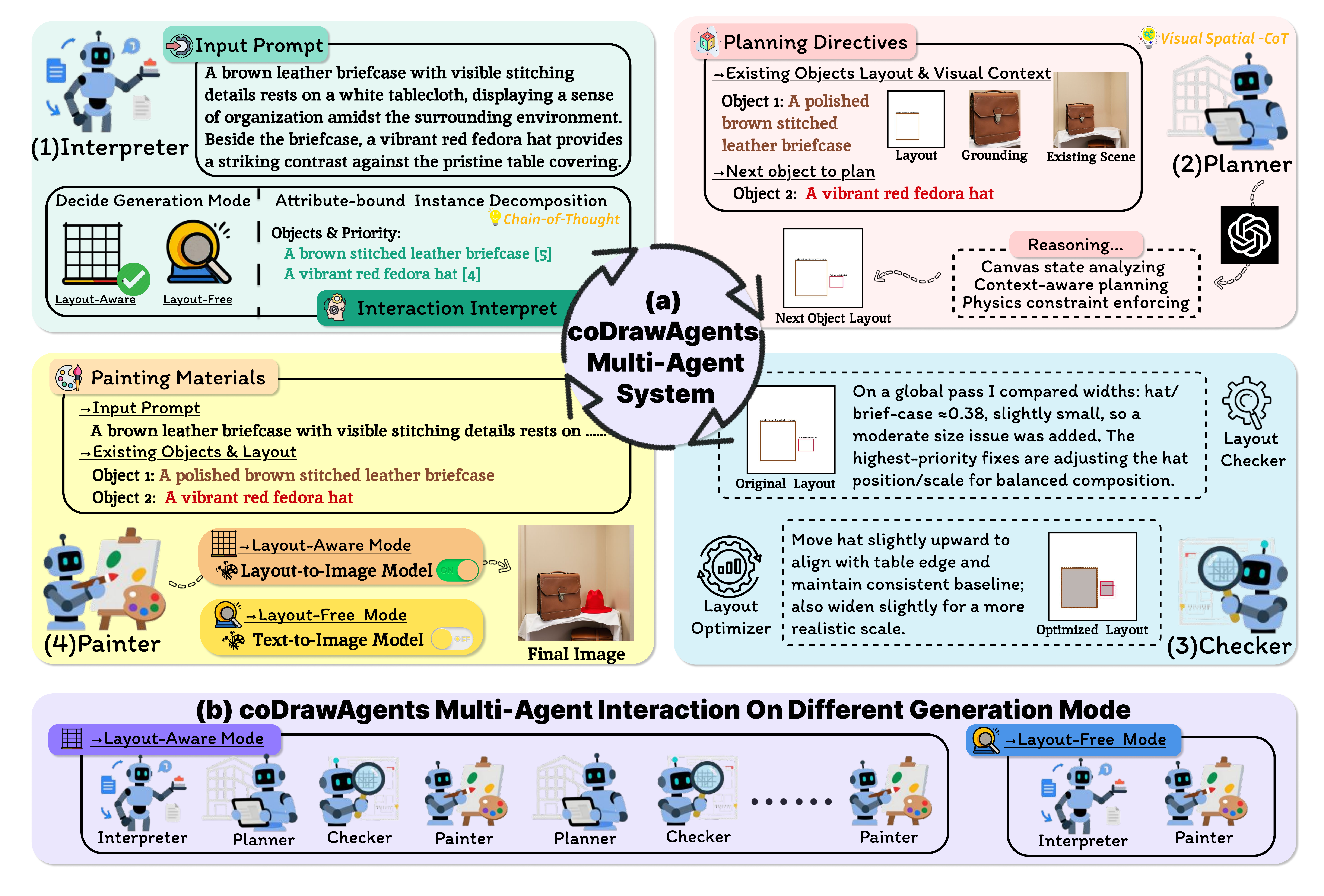}
\end{center}
\vspace{-20pt}
\caption{
Overview of our proposed coDrawAgents framework. 
(a) The multi-agent system consists of four specialized agents: (1) Interpreter, (2) Planner, (3) Checker, and (4) Painter. 
The Interpreter adaptively selects between layout-free and layout-aware modes; in the latter, the Planner incrementally proposes layouts, the Checker validates and refines them, and the Painter synthesizes the evolving canvas. 
(b) Illustration of the multi-agent interaction process in different generation modes, showing iterative collaboration in layout-aware mode and the direct pathway in layout-free mode.
}
\label{fig:overview}
\vspace{-12pt}
\end{figure*}

\vspace{-1pt}
\section{Related Work}
\label{sec:related_work}
\paragraph{Text-to-Image Generation}

The field of text-to-image (T2I) generation has seen rapid progress, initially driven by Vector Quantized GANs (VQGANs)~\citep{esser2021taming} paired with CLIP guidance~\citep{radford2021learning}. The paradigm shifted significantly with the advent of diffusion models, which led to remarkable improvements in image quality and image-prompt alignment. Foundational models such as DALL-E~\citep{ramesh2021zero}, Imagen~\citep{saharia2022photorealistic}, and Stable Diffusion~\citep{rombach2022high} established the potential of this new approach. Subsequent efforts have focused on scaling, with models including~\citep{betker2023improving,podell2024sdxl,FLUX.1-dev_github_2024,duong2025onediffusion} further enhance performance. More recently, there is a growing trend of integrating Multimodal Large Language Models (MLLMs) directly into the generation process to improve prompt comprehension and contextual reasoning~\citep{wang2024emu3,wu2024janus,yang2024cross,hu2024ella,gani2023llm,ding2021cogview,sun2023dreamsync,lee2024parrot,Janus-pro_arxiv_2025,Qu2025TOKENFLOW,UniWorld-V1_arxiv_2025,OmniGen2_arxiv_2025,gptimage,gao2025seedream,qwenimg}. However, these monolithic models still often struggle with fine-grained control over object composition and complex spatial relationships. 

\vspace{-13pt}
\paragraph{Layout-to-Image Generation}
To address the challenge of precise object placement, Layout-to-Image (L2I) generation conditions synthesis on explicit spatial information~\citep{zheng2023layout,feng2023layoutgpt,zhou20243dis,zhou20253disflux,nuyts2024explicitly,dahary2024yourself,jia2024ssmg,ma2024hico,lv2024place,zhang2025creatilayout}, typically in the form of bounding boxes or segmentation masks. 
ControlNet~\citep{zhang2023adding} and GLIGEN~\citep{li2023gligen} demonstrated spatial grounding in pre-trained diffusion models, while later works explored LLM-based layout generation~\citep{lian2023llm,feng2023layoutgpt}, training-free constraints~\citep{xie2023boxdiff}, and fine-grained regional controls~\citep{cheng2024rethink}.
These models require explicit spatial conditioning as an input, limiting their applicability when given only text input.

\vspace{-13pt}
\paragraph{Compositional Text-to-Image Generation}
Ensuring compositional faithfulness, where the generated images reflect all objects, attributes, and relations in a prompt, remains a key challenge. Early works such as Composable Diffusion~\citep{liu2022compositional} and Attend-and-Excite~\citep{chefer2023attend} combine concepts or refine attention guidance. Subsequent methods introduced layout reasoning as an intermediate step, for instance LayoutLLM-T2I~\citep{qu2023layoutllm}, LLM Blueprint~\cite{gani2023llm}, ALR-GAN~\citep{tan2023alr}, and LMD~\citep{lian2023llm}, which use LLMs or refinement modules to predict layouts that guide diffusion. RPG~\citep{yang2024mastering} extends this idea by denoising subregions in parallel, while PlanGen~\citep{plangen2024} integrates layout planning with synthesis. Most recently, GoT~\citep{fang2025got} employs a "Generation Chain-of-Thought" to produce a reasoning trace of semantic and spatial relations. While these approaches improve relational reasoning, they typically perform planning without visual feedback, making it difficult to resolve occlusion, depth, or other complex spatial interactions.

\vspace{-19pt}
\paragraph{Agent for Image Generation}
Recent works have begun to explore agent-based paradigms for image generation, ranging from multi-agent prompt decomposition~\citep{li2025mccd}, foreground-conditioned inpainting~\citep{tianyidan2025anywhere}, and self-correcting or interactive editing~\citep{wang2024genartist,wu2024self,ma2025talk2image}, to more recent directions such as self-improving agents~\citep{wan2025maestro}, multicultural generation~\citep{bhalerao2025multi}, training-free pipelines~\citep{chen2025t2i}, and proactive multi-turn dialogue~\citep{hahn2024proactive}. While these systems demonstrate the potential of agent designs, they are often limited by either fixed sequential pipelines (e.g., T2I-Copilot), planning solely from text without grounding (e.g., MCCD), or relying mainly on task scheduling or user queries without iterative visual feedback (e.g., Talk2Image, Proactive Agents). In contrast, our coDrawAgents framework introduces a closed-loop multi-agent dialogue where the Planner, Checker, and Painter interact continuously with the evolving canvas, achieving stronger interactivity and more faithful compositional generation.


\begin{figure*}[t!]
\begin{center}
\includegraphics[width=0.86\linewidth]{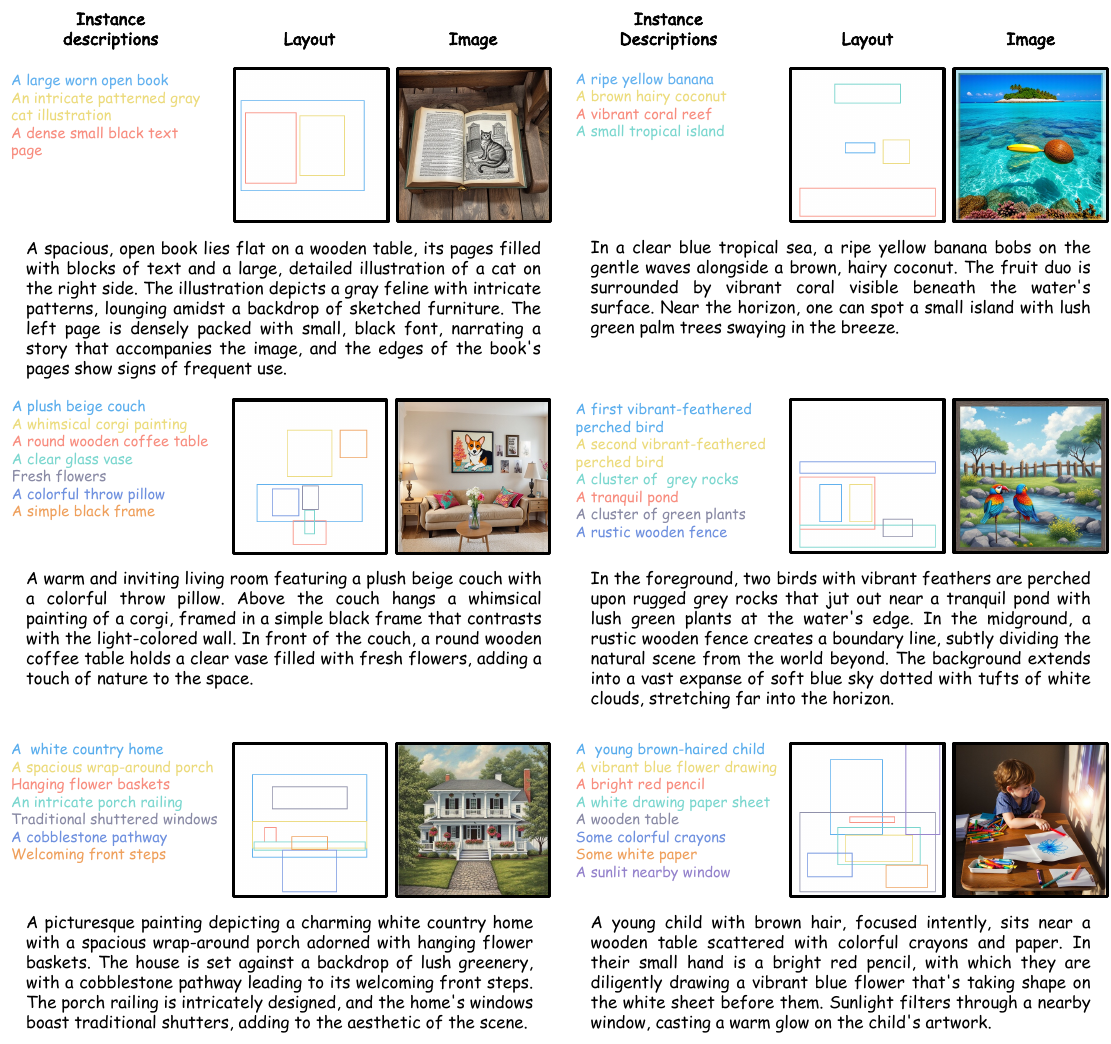}
\end{center}
\vspace{-15pt}
\caption{Generative results of our coDrawAgents framework.}
\label{fig:ours}
\vspace{-10pt}
\end{figure*}

\vspace{-4.5pt}
\section{coDrawAgents System}
\label{sec:method}
Given an input text $T$, our objective is to generate an image $I$ that faithfully aligns with the semantic content and spatial arrangement of the text. As shown in Figure~\ref{fig:overview}, we propose coDrawAgents, an interactive multi-agent dialogue framework in which four specialized agents: Interpreter, Planner, Checker, and Painter collaborate for compositional text-to-image generation. 
We will first introduce the details of agents collaboration and then each agent respectively in the remaining part of this section.

\vspace{-2pt}
\subsection{Multi-agent Collaboration}
To improve the generality of our approach, covering both general text-to-image cases without explicit layouts and more complex cases requiring layout planning, the Interpreter decides whether to enter the \emph{layout-free} or the \emph{layout-aware} mode.
In the \emph{layout-free} mode, the Interpreter directly invokes the Painter (a text-to-image model) to generate $I$ that aligns with $T$. 
In the \emph{layout-aware} mode, the Interpreter first parses $T$ into attribute-rich object descriptors, ranks them by semantic importance, and groups objects of similar priority for joint generation. The generation then proceeds through a Planner–Checker–Painter loop, iterating once for each semantic priority level.
In the \textit{$i^{th}$} iteration, Planner incrementally proposes layouts $L_i$ for the set of highest-salience objects at a time based on the existing objects grounding and scene visual context rather than the entire scene.
The Checker then leverages both the text and the visual context to validate spatial consistency and objects semantic alignment, followed by layout refinement.
The Painter synthesizes the image step by step, incorporating each newly planned object into the evolving canvas, which in turn provides essential visual context for subsequent iterations. After $N$ layout iterations, the final image $I$ is produced, closely aligned with the input text $T$.


In the layout-aware mode, the Planner adopts a divide-and-conquer strategy by reasoning about objects of the same semantic priority level at a time, which reduces layout complexity. Guided by the evolving canvas from the Painter, it grounds layout predictions in the actual scene rather than imagining them in isolation. The Checker further validates object placement and attribute alignment and refines layouts. This collaborative loop alleviates the burden of spatial planning and yields images that faithfully reflect the input text, particularly in complex object arrangements.

\subsection{Interpreter, Planner, Checker, and Painter}

In this section, we detail the four specialized agents: interpreter, planner, checker, and painter, each of whom is responsible for a distinct role in our generation process. The details of prompts are displayed in Supplementary Material.

\vspace{-4mm}

\paragraph{Interpreter}

To accurately represent complex scenes as a structured input for our iterative framework, we introduce an Interpreter agent to process $T$ for the agent system. The Interpreter first infers the relative importance of objects from $T$, and then decides whether to invoke the Painter directly for detail fidelity or to activate the multi-agent dialogue for layout-precise generation.


As illustrated in Figure~\ref{fig:overview}(a), in the layout-aware mode the Interpreter decomposes the prompt $T$ into structured, semantically rich object descriptors and prepares them for downstream planning. Specifically, we leverage large language models (LLMs) with chain-of-thought (CoT) prompting, guided by task instructions. The process follows three steps: (i) \textbf{Identify and decompose} the prompt into distinct semantic units; (ii) \textbf{Establish priorities} by ranking objects according to their semantic salience and grouping items with the same semantic-level for joint generation; and (iii) \textbf{Enrich attributes and background} through CoT-guided queries, yielding detailed descriptors of objects and their relations. The Interpreter then assigns the highest-priority objects of the current round to the iteration, enabling interactive generation with the Planner, Checker, and Painter in the generation loop.

\vspace{-4mm}

\paragraph{Planner}

The workflow of Planner is shown in Figure~\ref{fig:overview} (b). At $i^{th}$ iteration, Planner aims to plan the layout $L_i$ of the objects at the $i^{th}$ priority ranked by the Interpreter. Motivated by the multimodal chain-of-thought~\citep{zhang2024multimodal}, we propose a stepwise visualization chain-of-thought (VCoT) for layout planning. We employ GPT-5 as MLLM for VCoT.

VCoT takes as input the global text prompt $T$, the description of the $i^{th}$ priority objects, the layouts generated in the previous $i-1$ iterations, and the partial image $I_{i-1}$ rendered by Painter. It also incorporates object grounding, establishing correspondences between textual entities and image regions in $I_{i-1}$, which mitigates the inherent insensitivity of LLMs to spatial coordinates~\citep{you2024ferret} and enables reliable object localization.

We formulate our CoT reasoning as three steps: Canvas state analysis, Context-aware planning, and Physics constrain enforcement.  
In the ``Canvas state analysis'' stage, 
guided by the rich visual context of objects grounding, the image $I_{i-1}$ and other inputs, Planner meticulously analyzes the spatial layout of existing objects to gain a comprehensive visual understanding the current state of the scene. 
Afterwards, in the ``Context-aware planning'' stage, based on the existing canvas state, the MLLM planner leverages its embedded world knowledge to reason about the plausible interactions between the candidate object $O_i$ and the existing scene composition $(O_0, \dots, O_{i-1})$. 
Further, to maintain physical plausibility and scene coherence, 
Planner incorporate a ``Physics constrain enforcement'' module to prompt the MLLM to take physical and contextual constraints into account, which encourages realistic object placement to reflect real-world interactions and prevents issues like floating objects or improbable contacts. Please refer to Supplementary Material for more details on VCoT.

\vspace{-4mm}

\paragraph{Checker}
At each iteration $i$, the Checker performs a two-stage check–then–refine procedure illustrated in Figure~\ref{fig:overview}(c). 

\textbf{In the first stage}, it analyzes the current proposal $L_i$ and conducts checking at the object level (size, scale, boundary coverage) and the global level (relative placement among semantically related objects, inter-object relations, and overall spatial plausibility).  Based on these assessments, the Checker updates $L_i$ accordingly.

\textbf{In the second stage}, the Checker reviews all previous layouts $\{L_1,\ldots,L_i\}$ to identify cross-object conflicts such as overlaps, occlusion ordering inconsistencies, or scale drift. Guided by the global prompt, it examines and refines the layouts of all generated objects across iterations, allowing corrections of misplacements or other errors from earlier steps. For each identified issue, the Checker applies targeted, step-by-step fixes and propagates these corrections to subsequent layouts. The refined layout is then passed to the Painter for rendering.

\vspace{-4mm}

\paragraph{Painter}
The Painter supports two modes. In layout-free mode, it invokes a text-to-image (T2I) model to synthesize the image \(I\) directly from the prompt. In layout-aware mode, it uses a layout-to-image (L2I) model conditioned on current layout. Across iterations \(i\), the Painter incrementally renders the canvas by integrating each newly confirmed object,  providing visual context for subsequent steps. At the final iteration, the Painter renders the final image \(I\). 

The models used by Painter is designed to be plug-and-play, allowing any T2I and L2I model to be seamlessly integrated without additional training. 
In this paper, we use Flux~\cite{FLUX.1-dev_github_2024} for T2I model and 3DIS~\cite{zhou20253disflux} for L2I model. Note that our model is designed to be compatible with other, potentially more advanced L2I models, which could further improve our text-to-image generation performance.

\begin{table*}[t]
\centering
\caption{Performance comparison on the GenEval~\citep{Benchmark_geneval_nips_2023}. Best results are marked in \textbf{bold}. Column names are abbreviated to fit the page.}
\resizebox{\textwidth}{!}{
\begin{tabular}{lccccccc}
\toprule
\textbf{Model} & \textbf{Single Obj.} & \textbf{Two Obj.} & \textbf{Counting} & \textbf{Colors} & \textbf{Position} & \textbf{Color Attri.} & \textbf{Overall$\uparrow$} \\
\midrule

PixArt-$\Sigma$~\citep{Pixart_ECCV_2024} & 0.98 & 0.50 & 0.44 & 0.80 & 0.08 & 0.07& 0.48 \\ 
Emu3-Gen~\citep{Emu3-Gen_arxiv_2024} & 0.98 & 0.71 & 0.34 & 0.81 & 0.17 & 0.21 & 0.54\\
SDXL~\citep{sdxl_arxiv_2023} & 0.98 & 0.74 & 0.39 & 0.85 & 0.15 & 0.23 & 0.55 \\
GoT~\cite{fang2025got} &	0.99&	0.69&	0.67&	0.85&	0.34&	0.27 & 0.64 \\
DALL-E 3~\citep{DALL-E_3_arxiv_2020} & 0.96 & 0.87 & 0.47 & 0.83 & 0.43 & 0.45 & 0.67 \\
FLUX.1-dev~\citep{FLUX.1-dev_github_2024} &0.99 &0.81 &0.79 &0.74 &0.20 &0.47 &0.67 \\
Janus-Pro-1B~\citep{Janus-pro_arxiv_2025} &0.98 &0.82 &0.51 &0.89 &0.65 &0.56 &0.73 \\
SD3-Medium~\citep{SD3-Medium_ICML_2024} & 0.99 & 0.94 & 0.72 & 0.89 & 0.33 & 0.60 & 0.74 \\

TokenFlow-XL~\citep{Qu2025TOKENFLOW} &0.95 &0.60 &0.41 &0.81 &0.16 &0.24 &0.55\\
UniWorld-V1~\citep{UniWorld-V1_arxiv_2025} &0.99 &0.93 &0.79 &0.89 &0.49 &0.70 &0.80\\
GPT Image 1 [High]~\citep{gptimage}           & 0.99          & 0.92       & 0.85     & 0.92   & 0.75     & 0.61              & 0.84     \\

\textbf{coDrawAgents(Ours)}&\textbf{1.00}&\textbf{0.96}&\textbf{0.94}&\textbf{0.97}&\textbf{0.95}&\textbf{0.81}&\textbf{0.94}\\
\hline
\label{tab:gen_model_comparison}
\end{tabular}}
\end{table*}

\begin{figure*}[t!]
\begin{center}
\includegraphics[width=0.88\linewidth]{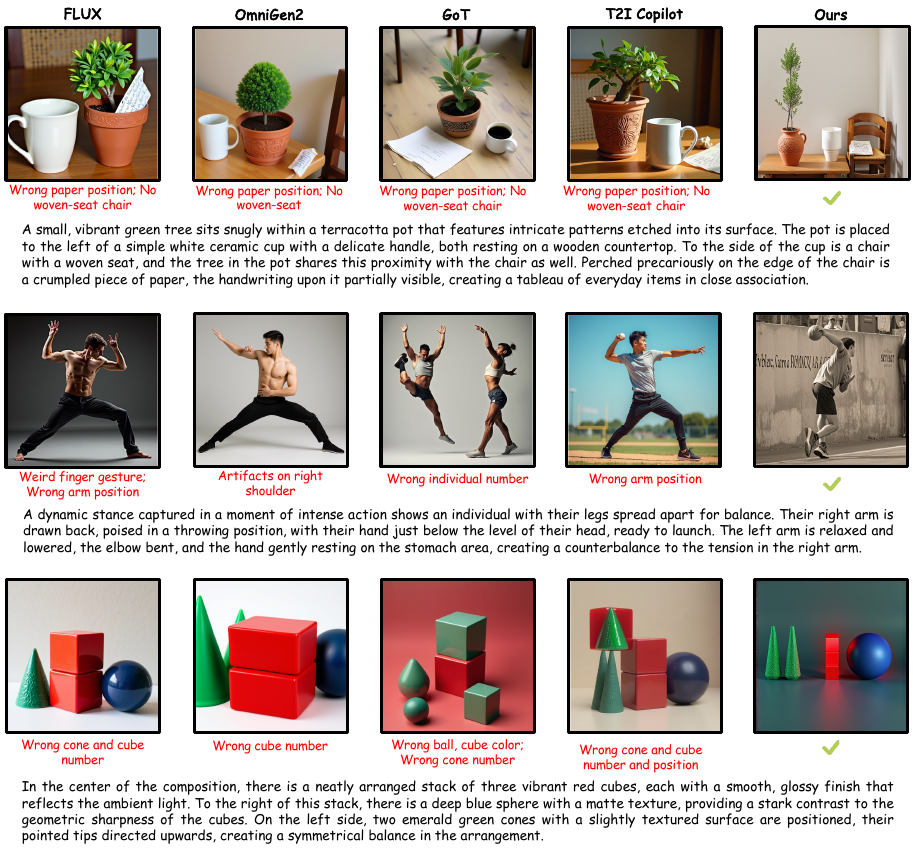}
\end{center}
\vspace{-15pt}
\caption{Qualitative comparison with existing methods.}
\label{fig: qual-comp}
\vspace{-10pt}
\end{figure*}

\begin{table*}[t!]
\centering
\caption{
Performance comparison on the DPG-Bench~\citep{Benchmark_DPG-Bench_arxiv_2024}. Best results are in \textbf{bold}.
}
\resizebox{\textwidth}{!}{%
\tiny
\begin{tabular}{l ccccc c}
\toprule
\noalign{\vskip -2pt}
\textbf{Model} & \textbf{Global} & \textbf{Entity} & \textbf{Attribute} & \textbf{Relation} & \textbf{Other} & \textbf{Overall$\uparrow$} \\
\noalign{\vskip -2pt}
\midrule
Hunyuan-DiT~\citep{Hunyuan-DiT_arxiv_2024} & 84.59 & 80.59 & 88.01 & 74.36 & 86.41 & 78.87 \\
PixArt-$\Sigma$~\citep{Pixart_ECCV_2024} & 86.89 & 82.89 & 88.94 & 86.59 & 87.68 & 80.54 \\
DALL-E 3~\citep{DALL-E_3_arxiv_2020} & 90.97 & 89.61 & 88.39 & 90.58 & 89.83 & 83.50 \\
SD3-Medium~\citep{SD3-Medium_ICML_2024} & 87.90 & 91.01 & 88.83 & 80.70 & 88.68 & 84.08 \\
FLUX.1-dev~\citep{FLUX.1-dev_github_2024}&74.35&90.00&88.96&90.87&88.33&83.84 \\
GoT~\citep{fang2025got} &	83.58&	82.16&	80.07&	87.81&	 65.25&  73.53 \\
T2I-Copilot~\citep{chen2025t2i} & 87.50 & 81.74 & 81.07 & 86.94 & 48.28 & 74.34 \\
OmniGen2~\citep{OmniGen2_arxiv_2025} & 88.81 & 88.83 & 90.18 & 89.37 & 90.27 & 83.57 \\
Emu3-Gen~\citep{Emu3-Gen_arxiv_2024} & 85.21 & 86.68 & 86.84 & 90.22 & 83.15 & 80.60 \\
UniWorld-V1~\citep{lin2025uniworld} & 83.64 & 88.39 & 88.44 & 89.27 & 87.22 & 81.38 \\
BLIP3-o 8B~\citep{BLIP3-O_arxiv_2025} & - & - & - & - & - & 81.60 \\
\textbf{coDrawAgents(Ours)}&84.78&90.15&87.55&92.92&84.38&\textbf{85.17}\\
\bottomrule
\end{tabular}%
}
\label{tab:dpg_bench_eval_vertical_resized}
\vspace{-11pt} 
\end{table*}

\vspace{-4.5pt}
\section{Experiments}
\label{sec:experiment}
\subsection{Dataset and Metrics}

We rigorously evaluated our coDrawAgents framework using two benchmark datasets: GenEval~\citep{ghosh2023geneval} and DPG-Bench~\citep{Benchmark_DPG-Bench_arxiv_2024}. GenEval, a standard for assessing text-to-image generation quality and text-image alignment, provides six metrics including object presence, attribute binding, counting, and spatial relationships.  We report the overall GenEval Score, along with its sub-scores, to quantify our model’s performance in each of these aspects. For DPG-Bench, which is designed to evaluate a model's ability to follow lengthy and dense prompts describing multiple objects with diverse attributes and relationships, we follow its established protocol, using an MLLM to adjudicate the generated images based on a series of questions. We report the overall DPG-Bench score, which is the average score across all prompts, along with scores for its main sub-categories. 

\subsection{Efficiency of coDrawAgents}
\label{sec:efficency_agent}
We report agent-usage statistics on DPG-Bench with 1,074 images. As shown in Table~\ref{tab:efficency_agent}, the Interpreter, Planner, Checker, and Painter are each invoked only a few times per generation (1.00, 1.52, 1.62, and 1.95 on average, respectively), which is far fewer than the average number of objects present in a scene (2.79). This efficiency arises because the Interpreter groups objects of the same semantic level, enabling multiple objects to be processed within a single round. This design substantially improves efficiency while maintaining strong performance.

\begin{table*}[t]
\caption{The average number of agent calls and objects in each DPG-bench image.} 
\label{tab:efficency_agent}
\centering
\renewcommand{\arraystretch}{1.6} 
\begin{tabular*}{\textwidth}{@{\extracolsep{\fill}} lcccc|c}
\toprule
\noalign{\vskip -5pt}
\textbf{Agent} & \textbf{Interpreter} & \textbf{Planner} & \textbf{Checker} & \textbf{Painter}  & \textbf{Number of Objects} \\
\midrule
\noalign{\vskip -5pt}
Avg.\ calls / generation & 1.00 & 1.52 & 1.62 & 1.95 & 2.79 \\
\noalign{\vskip -4pt}
\bottomrule
\end{tabular*}
\end{table*}

\begin{table*}[t!]
\centering
\tiny
\caption{
Quantitative ablation study. 
}
\resizebox{\textwidth}{!}{%
    \begin{tabular}{l ccccc c}
    \toprule
    \noalign{\vskip -2pt}
    \textbf{Model} & \textbf{Global} & \textbf{Entity} & \textbf{Attribute} & \textbf{Relation} & \textbf{Other} & \textbf{Overall$\uparrow$} \\
    \noalign{\vskip -2pt}
    \midrule
        Layout-free mode      & 84.50 & 84.44 & 86.15 & 90.87 & 75.60 & 77.60 \\
        + Layout-aware mode  & 79.94 & 89.32 & 87.27 & 92.37 & 80.65 & 82.61 \\
        + Visual context        & 88.89 & 88.72 & 89.32 & 95.95 & 66.67 & 84.51 \\
        + Checker (coDrawAgents)     & 84.78 & 90.15 & 87.55 & 92.92 & 84.38 & \textbf{85.17}\\

    \bottomrule
    \end{tabular}%
}
\label{tab:ablation}
\end{table*}

\begin{figure*}[t!]
\begin{center}
\includegraphics[width=0.8\linewidth]{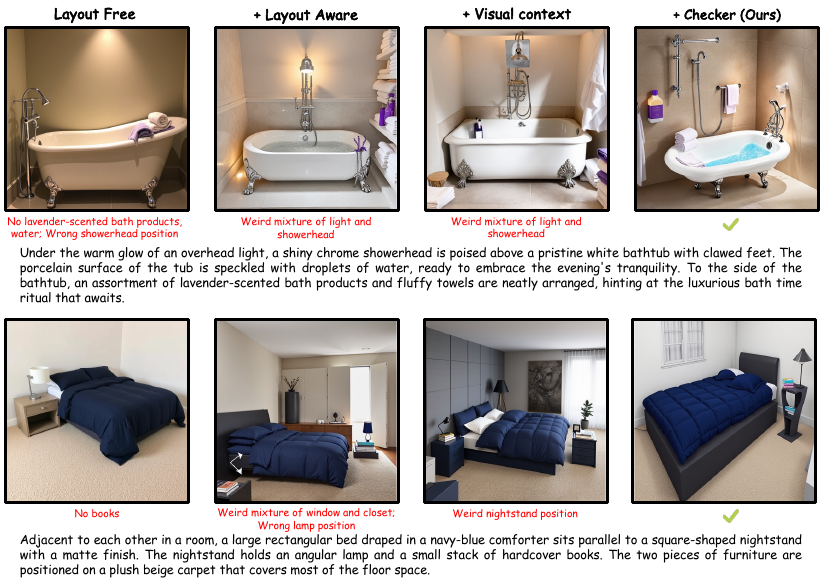}
\end{center}
\vspace{-15pt}
\caption{Qualitative ablation results. }
\label{fig: abl-vis}
\vspace{-10pt}
\end{figure*}

\subsection{Results and Comparisons}

As shown in Figure~\ref{fig:ours}, our coDrawAgents framework is capable of handling a wide range of challenging scenarios, including multi-object compositions, diverse visual styles, highly complex layouts with object interactions, and long descriptive prompts. The generated images remain highly consistent with the input descriptions while maintaining high visual quality. Additional results on DPG-Bench and GenEval, as well as our generative results on COCO~\cite{coco} and customized prompts demonstrating the framework's ability to generalize to complex scenes and inter-object relationships, are provided in Supplementary Material.

For quantitative evaluation, we compare coDrawAgents with a broad set of recent state-of-the-art models on GenEval and DPG-Bench (Tables~\ref{tab:gen_model_comparison} and~\ref{tab:dpg_bench_eval_vertical_resized}). 
The baseline models encompass a broad range of state-of-the-art text-to-image generation approaches, including representative T2I models such as DALL-E 3~\citep{DALL-E_3_arxiv_2020} and FLUX~\citep{FLUX.1-dev_github_2024}, recent multimodal large models like GPT Image 1 [High]~\citep{gptimage} and OmniGen2~\citep{OmniGen2_arxiv_2025}, layout-aware methods such as GoT~\citep{fang2025got} that perform explicit bounding box reasoning, and multi-agent frameworks like T2I-Copilot~\citep{chen2025t2i}. These models collectively represent diverse paradigms in the field, from direct generation to structured reasoning and interactive agent collaboration.
The coDrawAgents achieves the best overall results, demonstrating the effectiveness of our interactive multi-agent design.

For qualitative comparisons, we further compare coDrawAgents with representative methods from three categories: general text-to-image generation FLUX~\citep{FLUX.1-dev_github_2024}, vision-language models OmniGen2~\citep{OmniGen2_arxiv_2025}, 
one of the most recent open-source compositional image generation frameworks, GoT~\citep{fang2025got}, which explicitly performs one-shot reasoning over all bounding boxes during generation,
and the most recent multi-agent text-to-image generation framework T2I-Copilot~\citep{chen2025t2i}. As shown in Figure~\ref{fig: qual-comp}, existing approaches often suffer from misplaced objects, incorrect counts, or attribute artifacts, while our method produces coherent layouts and faithful compositions closely aligned with the textual descriptions.

\subsection{Ablation Studies}

We conduct ablation experiments on DPG-Bench to evaluate the contribution of each component in coDrawAgents. As shown in Table~\ref{tab:ablation} and Figure~\ref{fig: abl-vis}, starting from the layout-free baseline, introducing layout-aware planning allows LLM to explicitly generate layout plans and adopt a divide-and-conquer strategy over multiple objects rather than attempting global planning at once, thereby reducing reasoning complexity. Adding visual context enables Planner to leverage partially generated scene as grounding when placing the next set of semantically prioritized objects, which enhances spatial coherence. The Checker provides explicit error correction by detecting misplacements and attribute mismatches, further improving entity and attribute faithfulness. Our coDrawAgents model achieves the best overall balance, producing visually coherent scenes with stronger alignment between objects, attributes, and relations.

\section{Limitations and Future works}
While coDrawAgents Dialogue demonstrates significant progress in compositional text-to-image generation, it still has several limitations:

First, the multi-agent system, while beneficial for quality and compositional accuracy, introduces a computational overhead. The framework requires more processing time compared to single-pass methods due to the multi-agent calls. However, empirical analysis~\ref{sec:efficency_agent} shows coDrawAgents Dialogue still achieves competitive inference efficiency, outperforming many existing methods despite its iterative nature. Further optimization of the multi-agent loop remains a key area for future work.

Second, the performance of our Painter is inherently dependent on the underlying T2I and L2I models. This dependency means that limitations of the base models, such as imperfect attribute rendering or biased visual priors, may propagate into coDrawAgents, e.g., ``a radish with black skin''. Conversely, it also indicates that coDrawAgents will naturally benefit from future advances in text-to-image and layout-to-image generation. 

Third, the Planner and Checker rely on multimodal LLMs for layout reasoning and error detection, making the system susceptible to LLM-specific issues such as hallucination and overconfidence in incorrect layouts. These limitations may lead to invalid object placements or missed corrections, especially in highly compositional or ambiguous prompts. Conversely, it also indicates that coDrawAgents will naturally benefit from future advances in more reliable and grounded LLMs with reduced hallucination tendencies.

Fourth, while our current framework focuses on 2D compositional synthesis, extending this multi-agent dialogue system to the 3D domain presents a promising avenue for future work. Recent advancements in 3D generative models~\cite{liu2023exim, hui2022neural, hu2023neural}, have demonstrated impressive capabilities in controllable 3D generation~\cite{hu2023clipxplore, hu2024_cnsedit, hu2026pegasus3dpersonalizationgeometry, yan2026comp} and analysis~\cite{du2025hierarchical}. However, although existing methods have achieved impressive results in single-object synthesis and holistic scene- or city-scale 3D generation~\cite{feng2025wonderverse, liu2026imagine}, fine-grained 3D scene generation with precise, layout-driven spatial control remains largely underexplored. Extending our layout-aware, multi-agent collaborative loop to guide emerging 3D generative models could help bridge this gap, opening up new possibilities for complex, controllable, and compositionally faithful 3D scene synthesis.

Finally, as with most iterative frameworks, coDrawAgents may be affected by error accumulation across iterations. For instance, small placement inaccuracies in early steps can propagate if not fully corrected by the Checker. Nevertheless, our design explicitly mitigates this risk by introducing verification and refinement mechanism, and we observe that the overall error accumulation is significantly lower than in single-pass generation pipelines, as shown in Tables~\ref{tab:gen_model_comparison} and~\ref{tab:dpg_bench_eval_vertical_resized}.

\section{Conclusion}

In this paper, we introduced coDrawAgents, an interactive multi-agent dialogue framework for compositional text-to-image generation. The coDrawAgents brings together four specialized agents: the Interpreter that decomposes the text prompt, the Planner that reasons about object layouts, the Checker that verifies spatial and semantic consistency, and the Painter that renders the final image. By coordinating these roles, coDrawAgents addresses key challenges in complex scene generation, including layout reasoning, grounding in evolving visual context, and explicit error correction.
Evaluations on GenEval and DPG-Bench show that coDrawAgents achieves state-of-the-art performance, with notable gains in text–image consistency, spatial accuracy, and attribute binding compared with existing approaches. 

\section*{Acknowledgement}
This work has been supported by National Natural Science Foundation of China (Category C) fund code 62506149, Lingnan University StartUp Grant fund code: 103684, and Faculty Research Grant fund code:106106 and 106119.

{
    \small
    \bibliographystyle{ieeenat_fullname}
    \bibliography{main}
}


\end{document}